\documentclass[11pt]{article}
\usepackage[table]{xcolor}
\usepackage{inlg2017}

\makeatletter
\newcommand{\@BIBLABEL}{\@emptybiblabel}
\newcommand{\@emptybiblabel}[1]{}
\makeatother

\usepackage{times}
\usepackage[utf8x]{inputenc}
\usepackage{latexsym}
\usepackage{booktabs}
\usepackage{url}
\usepackage{tikz}
\usepackage{enumitem}
\usepackage{todonotes}
\usepackage{expex}
\lingset{textoffset=1ex}
\lingset{labeloffset=-2.5ex}
\lingset{preambleoffset=1ex}
\lingset{interpartskip=1px}
\lingset{aboveexskip={1.4ex plus .4ex minus .4ex}}
\lingset{belowexskip={1.4ex plus .4ex minus .4ex}}
\usepackage{stfloats}
\usepackage{microtype}

\naaclfinalcopy

\title{Cross-linguistic differences and similarities in image descriptions}

\author{Emiel van Miltenburg\\
Vrije Universiteit Amsterdam\\
{\tt emiel.van.miltenburg@vu.nl} 
\And
Desmond Elliott\\
University of Edinburgh\\
{\tt d.elliott@ed.ac.uk} 
\AND 
Piek Vossen\\
Vrije Universiteit Amsterdam\\
{\tt piek.vossen@vu.nl}
}

\date{}

\begin{document}
\maketitle
\begin{abstract}
Automatic image description systems are commonly trained and evaluated on large image description datasets. Recently, researchers have started to collect such datasets for languages other than English. An unexplored question is how different these datasets are from English and, if there are any differences, what causes them to differ. This paper provides a cross-linguistic comparison of Dutch, English, and German image descriptions. We find that these descriptions are similar in many respects, but the familiarity of crowd workers with the subjects of the images has a noticeable influence on description specificity. 
\end{abstract}

\section{Introduction}

Vision and language researchers have started to collect image description corpora for languages other than English, e.g.\ Chinese \cite{li2016adding}, German \cite{elliott2016multi30k,hitschlerETAL:16,rajendran2016bridge}, Japanese \cite{miyazaki2016cross,yoshikawa2017stair}, French 
\cite{rajendran2016bridge}, and Turkish \cite{unal2016tasviret}. The main aim of those efforts is to develop image description systems for non-English languages and to explore the related problems of cross-lingual image description \cite{elliott2015multilingual,miyazaki2016cross} and machine translation in a visual context \cite{specia-EtAl:2016:WMT,hitschlerETAL:16}. We view these new corpora as sociological data that is in itself worth studying. Our research stems from the following question: \emph{To what extent do speakers of different languages differ in their descriptions of the same images?} Considering this question, we developed a (non-exhaustive) list of factors that may influence the descriptions provided by crowd workers. Understanding the effect of these factors will enable us to improve the data collection process, and help us appreciate the challenges of natural language generation in a visual context:

\begin{enumerate}[topsep=2px,itemsep=0px]
\item \textbf{Task design effects}: There are many possible approaches to collecting descriptions of images. Previous research has re-used the Flickr8K \cite{hodosh2013framing} template and methodology. \newcite{baltaretu-castroferreira:2016:INLG} showed that task design may influence the form of crowd-sourced descriptions.

\item \textbf{(Perceived) audience}: Speakers 
adapt the style of their messages to their audience \cite{bell_1984}. Knowing how the descriptions will be used may affect the style or quality of the corpora.

\item \textbf{Individual/Demographic factors}: Individual features, like the demographics or personal preferences of the workers, may explain part of the variation in the descriptions.

\item \textbf{Differences in (background) knowledge}: Workers can only provide as much information as they know. Besides educational factors, the background knowledge of a person can be influenced by where they currently live and where they have previously lived.
\item \textbf{Language differences}: Languages differ in how they package information, which may be reflected in the descriptions. This is close to, but separate from \emph{linguistic relativity} (see e.g.~\cite{deutscher2010through,mcwhorter2014language}).

\item \textbf{Cultural differences}: Culture may influence the descriptions on a group level by affecting the social perspective of a population.

\end{enumerate}

This paper focuses on the last three factors in a cross-linguistic corpus study of Dutch, German, and English image descriptions. Our work is a starting point for understanding the differences in descriptions between languages. The focus on the last three factors is a consequence of our corpus study: the first two factors require manipulating the experimental set-up, and the third factor requires data about the crowd workers that is not known (and should ideally also be controlled). As we will see in Sections \ref{sec:characterize} and \ref{sec:familiarity}, we \emph{can} make claims about the last three factors based on the workers' language and geolocations.

We believe that studying differences between languages shows us which phenomena are robust across languages and thus important to consider when implementing and deploying models. Also, differences between languages can inform us about the feasibility of approaches to image description in different languages by translating existing English data \cite{li2016adding,yoshikawa2017stair}.

Our analysis combines quantitative and qualitative studies of a trilingual corpus of described images. We use the Flickr30K \cite{young2014image} for English, Multi30K for German \cite{elliott2016multi30k}, and a new corpus of Dutch descriptions (Section \ref{sec:collection}). We build on earlier work that studies the semantic and pragmatic properties of English descriptions \cite{van2016stereotyping,miltenburg2016pragmatic}.  Those works study ethnicity marking, negation marking, and unwarranted inferences about the roles of people. The main finding of our analysis is that all of these properties are stable across Dutch, US English, and German (Section \ref{sec:characterize}). We also show how differences in background knowledge can affect description specificity (Section  \ref{sec:familiarity}). We make the Dutch corpus available online and we also release software to explore image description corpora with the descriptions in different languages side-by-side 
to encourage future work with different language families.\footnote{\label{fn:resource}See:  \url{https://github.com/cltl/DutchDescriptions}}

\section{Related work}
We review work on the theory about the image description process, and work on automatic image description in other languages.

\textbf{Describing an image.}
Erwin Panofsky's \cite{panofsky1939studies} hierarchy of meaning was originally intended as a guide for interpreting works of art. It has since been applied by \newcite{shatford1986analyzing} and \newcite{jaimes1999conceptual} in the context of indexing and searching for images in libraries. The hierarchy consists of three levels that build on each other.

\begin{enumerate}[noitemsep,parsep=0px, topsep=2px]
\item \textbf{Pre-iconography}: giving a factual description of the contents of an image, and an expressional indication of the mood it conveys.
\item \textbf{Iconography}: giving a more \emph{specific} description, informed by knowledge of the cultural context in which the image is situated.
\item \textbf{Iconology}: interpreting the image, establishing its cultural and intellectual significance.
\end{enumerate}

This hierarchy is useful to think about for descriptions of images \cite{hodosh2013framing}. As \newcite{panofsky1939studies} notes, these levels require more knowledge as we move up the hierarchy. If we apply this hierarchy to the image description domain, we can say that image description corpora typically cover the first two levels. An important factor in the `quality' of a description is the amount of \emph{cultural} or \emph{background} knowledge that informs the description. We will 
explore the influence of this factor in Section~\ref{sec:familiarity}.

\textbf{Descriptions in other languages.}
Work on image description in other languages generally focuses on system performance rather than cross-linguistic differences \cite{elliott2015multilingual,li2016adding,miyazaki2016cross}. Thus far, any differences have only been anecdotally described.

\newcite{li2016adding} collected Chinese descriptions of images in the Flickr8K corpus \cite{hodosh2013framing}. They highlight the differences between Chinese and English descriptions using a picture of a woman taking a photograph. The English annotators describe the woman as \emph{Asian}, whereas Chinese annotators describe her as \emph{middle-aged}. The authors note that ``Asian faces are probably too common to be visually salient from a Chinese point of view.'' 

\newcite{miyazaki2016cross} collected Japanese descriptions for a subset of the MS COCO dataset, which mostly contains pictures taken in (or by people from) Europe and the United States \cite{lin2014microsoft}. They note that in their pilot phase, the images appeared ``exotic'' to Japanese crowd workers, who would frequently use adjectives like \emph{foreign} and \emph{overseas}. The authors actively tried to combat this by modifying their guidelines to explicitly prevent crowd workers using these phrases, but the observation remains that perspective can strongly influence the nature of the descriptions.

In this paper we collect a new dataset of Dutch image descriptions, but our work differs from previous work in two ways: (i) we aim to provide a more systematic overview of the differences between descriptions in three languages, and therefore (ii) we do not empirically evaluate system performance in reproducing the descriptions.
\section{Collecting Dutch descriptions}\label{sec:selection}\label{sec:collection}

We used Crowdflower to annotate 2,014 images from the validation and test splits of the Flickr30K corpus \cite{young2014image} with five Dutch descriptions.

Following other work, our goal is to create a parallel corpus of image descriptions, using the images as pivots. This requires us to stay as close to the original task setup as possible, thus fixing the effect of Task Design factor. We base our task on the template used by \cite{hodosh2013framing} to collect English descriptions, and by \cite{elliott2016multi30k} for German descriptions. In this design, images are annotated in batches of five images. The task for our participants is to describe each of those images ``in one complete, but simple sentence.'' Before starting on the task, we ask participants to read the guidelines, and to study a picture with example descriptions ranging from \emph{very good} to \emph{very bad}. We include the instructions for our task in the supplementary materials.

\textbf{Participants.} Crowdflower does not offer the option to select Dutch participants based on their native language. Instead, we restricted our task to level 2 (experienced and reasonably accurate) workers in the Netherlands. 
We had to continuously monitor the task for ungrammatical descriptions in order to stop contributors from submitting low-quality responses.

\textbf{Other settings.} Following \cite{elliott2016multi30k}, we set a reward for \$0.25 per completed task (or \$0.05 per image), and required participants to spend at least 90 seconds on each task, resulting in a theoretical maximum wage of \$10 per hour. We initially limited the number of judgments to 250 descriptions per participant, but due to the small size of the crowd we increased this limit to 500.

\textbf{Results.}
A total of 72 participants provided 10,070 valid descriptions in 116 days, at a cost of \$821.40. We were surprised by the number of  participants who presumably used Google Translate to submit their responses. These are identifiable through their ungrammaticality, usually due to incorrectly inflected verbs. An example is given in (\nextx), with a literal translation and original English description (verified using Google Translate).

\pex[textoffset=2ex] Response generated with Google Translate.
\a \ljudge* Een paar kussen \hfill {\small (Description)}\\
`A couple of kisses' \hfill {\small (Translation)}\\
A couple kisses \hfill {\small (Original)}
\xe

Altogether, we had to remove 60 participants due to either submitting ungrammatical responses (60\%), Lorum Ipsum text (12\%), random combinations of characters (9\%), non-Dutch responses (6\%), or otherwise low-quality responses (13\%). 

We conclude that crowdsourcing is a feasible way to collect Dutch data, but it may still be faster to collect image descriptions in the lab (in terms of time to collect the data, not counting the time spent as an experimenter overseeing the task). For large-scale datasets, such as Flickr30K or MS COCO, the Dutch crowdsourcing population seems to be too small to collect descriptions for \emph{all} the images in a reasonable amount of time. 
This is a problem; with the current data-hungry technology, low-resource languages and languages with smaller pools of crowd workers are in danger of being left behind. 
For example, \newcite{Sprugnoli2016} note that for Flemish, an example of a \emph{small-pool language}, they ``were not able to get a sufficient response from the crowd to complete the offered transcription tasks.''

\section{Characterizing English, German, and Dutch image descriptions}\label{sec:characterize}

We now examine the descriptions between languages in more detail, focusing on the \emph{validation} subset of the Multi30K dataset (1,014 images, with 5,070 descriptions per language). 

\subsection{General statistics}

\begin{table}
\centering\small
\begin{tabular}{lrrrr}
\toprule
 & Tokens & $\sigma$ & Words & $\sigma$\\
\midrule
Dutch & 11.14 & 4.5 & 10.32 & 4.3\\
English & 13.60 &  5.6 & 12.48 & 5.3\\
German & 9.76 & 4.2 & 8.81 & 3.9\\
\bottomrule
\end{tabular}
\caption{Mean sentence length across languages.}
\label{table:mean-sentence-length}
\end{table}

Table~\ref{table:mean-sentence-length} shows the mean sentence length (in tokens and words) for the three languages. The English descriptions are the longest, followed by the Dutch and the German ones. However, German has the longest average word length (5.25 characters per word), followed by Dutch (4.62) and English (4.12). This difference seems due to German and Dutch compounding, which is confirmed by the number of word types: German has 31\% more types than English (5709 versus 4355). Dutch has 19\% more (5193).

\begin{table*}
\centering
\begin{minipage}{0.37\textwidth}
\raggedleft\small
\begin{tabular}{llr}
\toprule
Dutch & Gloss & Count\\
\midrule
Een man  & A man & 517\\
Een vrouw  & A woman & 252\\
De man  & The man & 105\\
Een jongen  & A boy & 92\\
Twee mannen  & Two men & 92\\
\bottomrule
\end{tabular}
\end{minipage}
\begin{minipage}{0.23\textwidth}
\centering\small
\begin{tabular}{lr}
\toprule
English & Count\\
\midrule
A man  & 760\\
A woman  & 367\\
A young  & 223\\
A group  & 211\\
Two men  & 127\\
\bottomrule
\end{tabular}
\end{minipage}
\begin{minipage}{0.37\textwidth}
\raggedright\small
\begin{tabular}{llr}
\toprule
German & Gloss & Count\\
\midrule
Ein Mann  & A man & 584\\
Eine Frau  & A woman & 296\\
Zwei M\"anner  & Two men & 120\\
Ein Junge  & A boy & 108\\
Der Mann  & The man & 93\\
\bottomrule
\end{tabular}
\end{minipage}

\caption{Top-5 most frequent bigrams at the start of a sentence, with their English translation.}\label{bigrams}
\end{table*}

\textbf{Definiteness.}
The five most frequent bigrams that start a description (showing the typical subjects of the images) are given in Table~\ref{bigrams}. The majority starts with an indefinite article, which is in line with the \emph{familiarity theory of definiteness}: the function of definite articles is to refer to familiar referents, whereas indefinite articles are used for unfamiliar referents \cite{christophersen1939articles,heim1982semantics}. The distribution of (in)definite articles follows from the fact that the participants have never seen the images before, nor any context for the image in which the referents could be introduced. A corollary is that systems trained on this data are more likely to produce indefinite than definite articles, and need to be told when definites should be used.

\subsection{Replicating previous findings for negation and ethnicity marking}

Previous work has studied the use of negation and ethnicity marking in English image description datasets \cite{miltenburg2016pragmatic,van2016stereotyping} We now attempt to replicate these findings in the Dutch and German data.

\textbf{Negations.} \newcite{miltenburg2016pragmatic} performed a corpus study to categorize all uses of (non-affixal) negations in the Flickr30K corpus. Negations are interesting in descriptions because they describe images by saying what is \emph{not} there. Negations may be used because something in the picture is unexpected, goes against some social norm, or because non-visible factors are relevant to describe the picture. If annotators consistently use negations, this can be seen as evidence that the negated information is part of their shared background knowledge and is a strong requirement for producing human-like descriptions. We readily found examples of negations in both the Dutch and the German data. Some examples are given in (\ref{ex:dutchnegations1}) and (\ref{ex:germannegations1}), respectively.

\lingset{aboveexskip={5px}}
\pex Examples from the Dutch descriptions\label{ex:dutchnegations1}
\a De kinderen dragen \textbf{geen} kleding.\\
`The kids are \textbf{not} wearing any clothing.'
\a Vrouw snijdt broodje \textbf{zonder} te kijken(!)\\
`Woman slices a bun \textbf{without} looking(!)'
\xe

\lingset{aboveexskip={0px}}
\pex Examples from the German descriptions\label{ex:germannegations1}
\a Zwei Buben \textbf{ohne} T-Shirt setzen auf der Stra\ss e.\\
`Two boys without T-shirt sitting on the street.'
\a Eine Ansammlung von Menschen [\ldots] schaut auf ein Ereignis, das \textbf{nicht} im Bild ist.\\
`A crowd of people is watching an event not shown in the picture.'
\xe
\lingset{aboveexskip={2ex plus .4ex minus .4ex}}

In total, we found 11 Dutch and 20 German descriptions containing explicit negations in the corpus, while \newcite{miltenburg2016pragmatic} found 27 in English for the same images (excluding false positives).  This confirms that workers in different languages mark negations at approximately the same rate, given a sample size of 5,070 sentences.
We found almost no images that consistently attracted the use of negations in all three languages: we found only four examples of co-occurring negation between languages. One image is described by speakers of all three languages using a negation (a man with two prosthetic legs, described as having no legs), and there are three other images (all of shirtless individuals) where both English and German workers use negations.

\textbf{Racial and ethnic marking.} \newcite{van2016stereotyping} found that the descriptions in the Flickr30K data have a skewed distribution of racial and ethnic markers: annotators used terms like \emph{asian} or \emph{black} much more often than  \emph{white} or \emph{caucasian}. If we find the same disproportionate use of ethnicity markers in Dutch and German, then we can conclude that this is not a quirk in the English data, but a systematic linguistic bias \cite{beukeboom2014mechanisms}.

Indeed, we did find that non-white people were often marked with adjectives such as \emph{black, dark-skinned, Asian, Chinese}. In Dutch and German, white people were only marked to indicate a contrast between them and someone of a different ethnicity in the same image. The English data contains five exceptions to this rule, where white individuals were marked without any people of another ethnicity being present in the image. We do have to note, however, that there are other ways to \emph{indirectly} mark someone as white, e.g.\ using adjectives like \emph{blonde} or \emph{brunette}. 

\begin{figure}
\centering
\begin{tikzpicture}[scale=0.8]
\draw[thick] (2,2) ellipse (2cm and 1cm); 
\draw[thick] (3,2) ellipse (2cm and 1cm); 
\draw[thick] (2.5,1.25) ellipse (2cm and 1cm); 

\draw (0.4,2.7) node[anchor=east] {\smash{\small Dutch}};
\draw (4.5,2.7) node[anchor=west] {\smash{\small German}};
\draw (0.9,0.4) node[anchor=east] {\smash{\small English}};

\draw (0.5,2.25) node {\small 17}; 
\draw (4.5,2.25) node {\small 18}; 
\draw (2.5,2.6) node {\small 12}; 
\draw (0.9,1.45) node {\small 11}; 
\draw (2.5,1.75) node {\small 35}; 
\draw (4.1,1.45) node {\small 33}; 
\draw (2.5,0.6) node {\small 15}; 

\end{tikzpicture}
\caption{Venn diagram of ethnicity markers by Dutch, English, and German workers. Counts correspond to images. }
\label{fig:race-counts}
\end{figure}
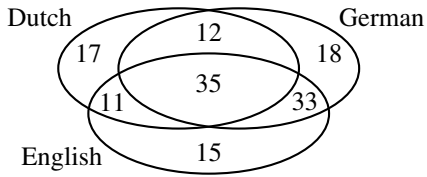

Figure~\ref{fig:race-counts} shows a Venn-diagram of the use of race/ethnicity markers in Dutch, English, and German. 
We observe that English and German workers use these markers slightly more often than  Dutch workers. However, we do \emph{not} claim that this is evidence that people living in Germany and the U.S.A. are \emph{more racist} than people living in the Netherlands. Rather than trying to interpret the meaning of this difference, we ask a different question: what drives people to mention racial or ethnic features?

There are several reasons why people may mark race/ethnicity in their descriptions. One common theme is that annotators mark images where the people are dressed in traditional outfits. Examples include traditional dancers from South-East Asia, and Scotsmen wearing kilts. These items of clothing are \emph{meant to} signal being part of a group, and the annotators picked up on this.

The distribution of the labels may be explained in terms of markedness \cite{jakobson1972verbal} and reporting bias \cite{misra2016seeing}. In this explanation, white is seen as the unmarked default, as it is the dominant ethnicity in all three countries.\footnote{
The US population is 75\% white, according to the 2010 census \cite{US-census}. The Dutch and German census bureaus do not monitor ethnicity, and instead report that 77\% of the Dutch population is Dutch/Frisian \cite{cbs} and 80\% of the German population is German \cite{destatis}.
} 
The marker \emph{white} is only used to be consistent in the use of modifiers within same sentence. This reasoning also explains the observation by \newcite{miyazaki2016cross} that Japanese crowd workers often used the labels \emph{foreign} and \emph{overseas} for the MS COCO images.

A final reason for crowd workers to mention ethnicity and skin color may be that the images are visually less interesting, but the description task still demands that the workers provide a description. Workers are thus pressured to find \emph{something} worth mentioning about the image, because too general descriptions might get their work rejected. This is a general task effect that may have implications beyond racial/ethnic marking.

\textbf{Speculation.} \newcite{van2016stereotyping} also found that that annotators often go beyond the content of the images in their descriptions, making \emph{unwarranted inferences} about the pictures. If we find that Dutch and German crowd workers also make such inferences, we conclude that image descriptions in all three languages are \emph{interpretations} of the images that may not necessarily be true.

\begin{figure}
\centering
\includegraphics[trim={140 50 0 80},clip,width=0.55\linewidth]{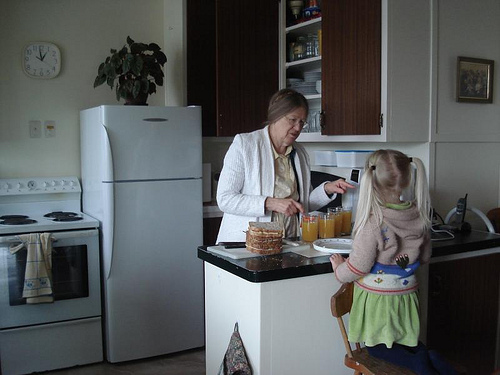}
\caption{Image 4634063005. The older woman in the picture was often seen as the grandmother.}
\label{fig:grandmother}
\end{figure}

We observed unwarranted inferences throughout the Dutch and German data, especially about women with infants, who were often seen as the mother. Figure~\ref{fig:grandmother} shows an image where both Dutch, English, and German workers suggested the woman is the \emph{grandmother}. In the most extreme case, two KLM stewards in pantsuits were described by a German worker as well-dressed \emph{Lesben} (`lesbians'). It would be undesirable for a model to associate all unseen images of air stewards with lesbians.
We expect that having multiple descriptions alleviates this type of extreme example, but there is an open question about how to deal with more common types of speculation.


\section{Familiarity and cultural differences}\label{sec:familiarity}

As the speakers of Dutch, English, and German have different backgrounds, some images may be more familiar to one group than to the others. Familiarity enables speakers to be more specific (but doesn't necessarily cause them to \emph{be} more specific). We will look at three kinds of examples (selected after inspecting the full validation set), where differences in familiarity lead to differences in the description of named entities, objects, and sports. These examples are illustrative of a larger issue, namely that descriptions in one language may not be adequate for speakers of another language (even if they were perfectly translated). We discuss this issue in \S\ref{sec:limits-of-translation}.

\subsection{Named entities}
The Dutch, English, and German descriptions differ in their use of place and entity names. 
We study two cases: one image that is more likely to be familiar to European workers (German and Dutch), and one that is more likely to be familiar to US workers (English).

\begin{figure}
\centering
\includegraphics[trim={0 30 0 70}, clip,width=0.5\linewidth]{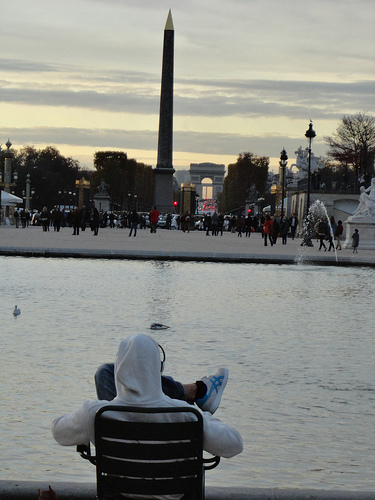}
\caption{Image 6408975653. This picture was taken at the Tuileries Garden in Paris, and shows the Luxor Obelisk and the Arc de Triomphe.}
\label{fig:tuileries}
\end{figure}

\textbf{The Tuileries Garden.} Figure~\ref{fig:tuileries} shows a scene from the Tuileries Garden in Paris, a popular tourist attraction. It may be more likely for a European crowd worker to have visited this location than for an American crowd worker. Three Dutch people indeed included references to the actual location in their description. One mentioned the Arc de Triomphe in the background, one said that this picture is from a square in Paris, and the most specific description (correctly) identified the location: 

\lingset{aboveexskip={5px}}
\ex
Een man zit aan de vijver van het Tuilleries park in Parijs.\\
`A man is sitting by the pond of the Tuileries park in Paris.'
\xe

Neither the German nor the American workers identified the location or the monuments by name (though one American worker thought this picture was taken at the Washington Monument).
Instead of mentioning the location, the English and German workers describe the scene in more general terms. Two examples are given in Example \nextx.

\lingset{labeloffset=0.8ex}
\pex<pg>
\a A person in a white sweatshirt is sitting in a chair near a pond and monument.

\a A man in a white hoodie relaxes in a chair by a fountain.
\xe
\lingset{labeloffset=-1.75ex}
\lingset{aboveexskip={2ex plus .4ex minus .4ex}}

These examples reveal a common strategy to handle unfamiliarity: focus on something else you \emph{do} know. This undermines the idea that crowd-sourced descriptions tell us what is relevant about the picture.

\begin{figure}
\centering
\includegraphics[trim={0 0 0 5}, clip, width=0.6\linewidth]{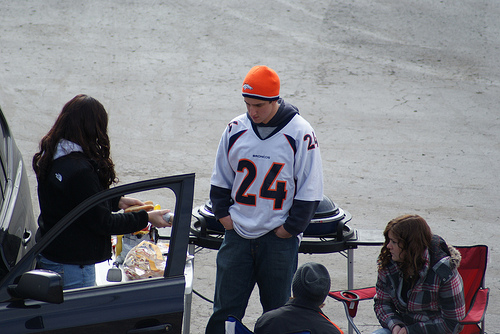}
\caption{Image 4727348655. This picture shows a man wearing a Denver Broncos hat and jersey.}
\label{fig:broncos}
\end{figure}

\textbf{The Denver Broncos.} Figure~\ref{fig:broncos} shows a man wearing a Denver Broncos hat and jersey. The Denver Broncos are an American Football team, which is not so well-known in Europe. Two American crowd workers but neither the Dutch nor the German workers identified the Broncos jersey. Three out of five American workers also described the activity in the image as \emph{tailgating}, a typical North-American phenomenon where people gather to enjoy an informal (often barbecue) meal on the parking lot outside a sports stadium. As this concept is not so prevalent in Dutch or German culture, there is no Dutch or German word, idiom, or collocation to describe tailgating. Such `untranslatable' concepts are called \emph{lexical gaps}. The presence of this gap means that the Dutch and German workers can only concretely describe the image without being able to relate the depicted event to any more abstract concept.

\subsection{Objects}
\begin{figure}
\centering
\includegraphics[trim={0 45 0 0}, clip, width=0.6\linewidth]{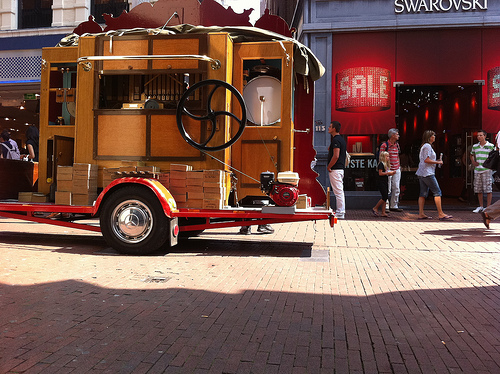}
\caption{Image 4897113571. This picture shows the back of a street organ in the Netherlands.}
\label{fig:draaiorgel}
\end{figure}

Familiarity also plays a role in labeling objects. Consider Figure \ref{fig:draaiorgel}, which shows (the backside of) a street organ in a shopping street in the Netherlands. All Dutch workers, as well as two German workers identified this object as a street organ, whereas the English workers are only able to provide very general descriptions (Example \nextx).

\lingset{labeloffset=0.8ex}
\pex
\a A \textbf{yellow truck} is standing on a busy street in front of the Swarovski store.
\a A \textbf{strange looking wood trailer} is parked in a street in front of stores.
\a An \textbf{unusual looking vehicle} parked in front of some stores.
\xe
\lingset{labeloffset=-1.75ex}

This example illustrates two strategies the crowd may use to provide descriptions for unfamiliar objects: (1) signal the unfamiliarity of the object using adjectives like \emph{strange} and \emph{unusual looking}. This is similar to the finding by \newcite{miyazaki2016cross} that the Japanese crowd made frequent use of terms like \emph{foreign} and \emph{overseas} for the Western images from MS COCO. (2) use a more general cover term, like \emph{vehicle}. Such terms may have a higher \emph{visual dispersion} \cite{kiela-EtAl:2014:P14-2}, but they provide a safe back-off strategy.

\subsection{Sports}

We found that unfamiliarity with different kinds of sports leads to the misclassification of those sports. We focus on three sports: American Football, Rugby, and Soccer. Looking at images for these sports, we compared how the three different groups referred to them. We found that the German and Dutch groups patterned together, deviating from the American crowd workers.

As expected, the Dutch and German workers make the most mistakes categorizing American Football. For all seven pictures of American Football, there is at least one Dutch annotator who thinks it's a game of Rugby. For six of those, at least one German annotator made the same mistake. By contrast, workers from the US made more mistakes identifying rugby images. For all three pictures of Rugby, there is at least one American calling it Soccer or Football. For one of those images, a German annotator thought it was American Football. All Soccer images were universally recognized as Soccer.

\section{Discussion}

\subsection{Description specificity}
In Section \ref{sec:familiarity} we observed that annotators differ in the specificity of their descriptions due to their familiarity with the depicted scenes or objects. One challenge for image description systems is to find the right level of specificity for their descriptions, despite this variation. If a system can identify the exact category of an object, it is probably more useful to produce e.g. \emph{street organ} rather than \emph{unusual looking vehicle}.

Besides familiarity, there are also other factors influencing label specificity.
For example, cultures may have differences in their \emph{basic level}; i.e.\ how specific speakers are generally expected to be \cite{rosch1976basic,matsumoto1995conversational}. For this reason, \emph{dog} is a more appropriate label than \emph{affenpinscher} in most situations, even though the latter is more specific. Ideally, image description systems should recognize when to use a more general term, and when to go more into detail \cite{entrylevelIJCV2015}.

\subsection{Limitations of translation approaches}\label{sec:limits-of-translation}
One approach to image description in multiple languages is to use a translation system. For example, \newcite{li2016adding} compare two strategies: \emph{early} versus \emph{late} translation. Using early translation, image descriptions are translated to the target language before training an image description system on the translated descriptions. Using late translation, an image description system is trained on the original data, and the output is translated. \newcite{li2016adding} show that the former strategy achieves the best result, and argue that it is a promising approach because it requires no extra manual annotation.

Our observations in Section~\ref{sec:familiarity} show that there are limits to what a translation-based approach can achieve. While translation provides a strong baseline, it can only capture those phenomena that are familiar to the crowd providing the descriptions. The street organ example shows that there exists a `knowledge gap' between Dutch and English. Dutch users would certainly not be satisfied with street organs being labeled as \emph{unusual looking vehicles}. If the translation-based approach is to be successful, future research should find out how to bridge such gaps.

\subsection{Limitations of this study}

Our focus on Germanic languages from the Western world does not allow us to make general statements about how people describe images. A comparison with taxonomically and culturally different languages might help us uncover important factors that we have missed in this study. A surprising example comes from \newcite{baltaretu2016talking}, who discuss how writing direction (left-to-right versus right-to-left) affects the way people process and recall visual scenes. This may have implications for the way that images are described by (or should be described for) speakers of languages differing in this regard.

Finally, there are limits to what a corpus study can show. The phenomena described here are presented with post-hoc explanations. Plausible as these explanations may be, they are still hypotheses. We think these hypotheses are useful guides in thinking about image description, but they still remain to be validated experimentally.

\section{Conclusion}
We studied a trilingually aligned corpus of described images to learn about how crowd workers of different languages described the same images. The main finding was that earlier observations about negation marking and ethnicity marking by English workers also hold for Dutch and German. Dutch and German workers also use negations in their image descriptions, showing that this is a robust phenomenon. Dutch and German workers also make unwarranted inferences about the images, this shows that crowd workers regularly include extra-visual information in their descriptions. In addition, Dutch and German workers also disproportionately mark non-white people in their descriptions, showing that image description corpora carry biases that we need to take into account when working with this data. 

We also explored the role of familiarity in image description. We found images in our corpus that were easily described by workers of one language, but unidentifiable to the workers of another language. This has consequences for image description models trained on automatically translated training data: some images will not be properly described for the target audience. But the problem is more general. The success of image description systems trained on datasets of described images is limited by the knowledge of the annotators, regardless of the language. While the available data is useful for us to learn and discuss what human-like descriptions should look like, it can only take us so far. Full coverage systems that could tailor their descriptions to particular audiences are still out of reach.

%


We hope this work provides a starting point for conducting cross-linguistic comparisons of image descriptions. Future work includes replicating our analyses across more diverse families of languages, modifying the task design to contrast the results with our findings, and using our inspection tool to explore other linguistic phenomena. We are also interested in scaling up our analyses to larger corpora, which will require the development of automated comparison methods. We believe that these steps will bring us closer to an initial understanding of the diversity in image descriptions across different languages and social groups.

\section*{Acknowledgements}
This research is funded through the NWO Spinoza prize, awarded to PV. DE is supported by an Amazon Research Award. We thank three anonymous reviewers for their questions and comments.


\bibliography{bibliography}
\bibliographystyle{inlg2017}

\end{document}